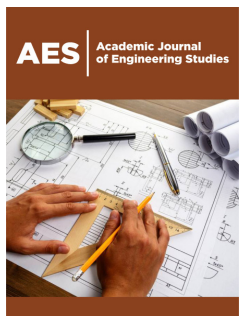

# A Study on Centralised and Decentralised Swarm Robotics Architecture for Part Delivery System

**Angelos Dimakos[1], Daniel Woodhall[1] and Seemal Asif[2]***

[1]School of Aerospace, Transport and Manufacturing, Cranfield University, UK

[2]Centre for Structures, Assembly and Intelligent Automation, School of Aerospace, Transport and Manufacturing, Cranfield University, UK

**Abstract**

Drones are also known as UAVs are originally designed for military purposes. With the technological advances, they can be seen in most of the aspects of life from filming to logistics. The increased use of drones made it sometimes essential to form a collaboration between them to perform the task efficiently in a defined process. This paper investigates the use of a combined centralised and decentralised architecture for the collaborative operation of drones in a parts delivery scenario to enable and expedite the operation of the factories of the future. The centralised and decentralised approaches were extensively researched, with experimentation being undertaken to determine the appropriateness of each approach for this use-case. Decentralised control was utilised to remove the need for excessive communication during the operation of the drones, resulting in smoother operations. Initial results suggested that the decentralised approach is more appropriate for this use-case. The individual functionalities necessary for the implementation of a decentralised architecture were proven and assessed, determining that a combination of multiple individual functionalities, namely VSLAM, dynamic collision avoidance and object tracking, would give an appropriate solution for use in an industrial setting. A final architecture for the parts delivery system was proposed for future work, using a combined centralised and decentralised approach to combat the limitations inherent in each architecture.

***Corresponding author:** Seemal Asif, Centre for Structures, Assembly and Intelligent Automation, School of Aerospace, Transport and Manufacturing, Cranfield University, UK

## Introduction

The rapidly increasing need for faster production speeds necessitates the application of drones for use in factories and industrial applications. Human-robot collaborative working is vital for the progression of the factories of the future, with robots facilitating and expediting the work of human operators. This article investigates the use of a decentralised system architecture for the collaborative intra-logistical operation of a swarm of drones, to be used as a parts delivery system in an industrial application. Experimentation was undertaken to determine the efficacy and plausibility of both the centralised and decentralised approaches, comparing the two to determine an optimal architecture for a swarm of UAVs in a parts delivery task. The Internet of Robotic Things provides a network between intelligent, autonomous agents. Shared perception and actuation enable increased productivity in a variety of industries. However, significant communication between different hardware is essential. The use of a centralised architecture can lead to a rapid increase in computational requirements and communication time resulting in instability and real-time performance issues; thus, the use of a decentralised framework was investigated. Swarm robotics can mitigate these issues, by imitating the behavioural pattern of animal flocks, and limiting the complexity of individual control, as described by [1].

Specifically, in the domain of intralogistics, synergy between humans and machines is essential. Robustness and safety are major concerns when having automation in proximity with personnel. As such, various experimental setups have been created to validate the efficiency of an autonomous delivery system in an indoor setting. An experimental setup for a swarm of drones in an intralogistics scenario has been investigated in the work of [2]. Here, the decentralised approach and the networking for a swarm of Unmanned Aerial Vehicles (UAVs) was studied. The study determined that a well-structured environment is essential, both for the construction of a global reference frame and to ensure smooth movement based on visual perception. The factories of the future have increasing demands for flexibility in their production lines. Drones can address this need through their increased available workspace and small size factor. The movement speed of agents is another factor that results in a significant reduction in the delivery time of parts. UAVs can address multiple needs such as the movement of components; [3]. However, the application of drones in the manufacturing sector at present is minimal. Limited hardware capabilities and safety concerns restrict their practical use. The low energy efficiency and limited payload capabilities hamper their potential benefits in real-world scenarios. To increase the adoption rate, a specific infrastructure is necessary. In terms of trajectory, only automatic and autonomous paths can be applied. This is to reduce the transfer of data and limit the use of sensors for collision avoidance. Scheduling of the operational time should take into consideration the charging time necessary for each drone [4]. Here, each drone picks up an object, limiting the weight of parts to below 2 kilograms. This technology can be utilised in reconfigurable manufacturing systems, which can better respond to unexpected changes, as described by [5]. Autonomous logistics are a fundamental consideration for smart factories. The automated delivery of parts can enable increased flexibility in existing flows, as human operators are relieved of tedious tasks [6]. The factories of the future should aim to expedite the production process as much as possible, to allow for rapid growth and development of industry. Through the incorporation of human-robot collaborative efforts into the production line, the overall process can be significantly streamlined and optimised, allowing for vastly increased rates of production. In an industrial setting where human operators are present, time for the transportation of sub-assemblies or acquisition of the necessary tools for production to progress can significantly decrease the overall speed of a production line. Through automating the retrieval of sub-assemblies and tools, the need for excessive movement by the human operator around the factory workspace is greatly reduced. The automation necessary for such a system could be achieved by employing a swarm of parts delivery drones to act as a human-robot collaborative system. A major concern with any human-robot collaborative operation is safety, with any system that is implemented needing to adhere to strict safety standards, with stringent programming features that ensure the safety of the human operator is held at the highest priority.

In this article, a parts delivery system is proposed that would assist in the production process, whilst incorporating collision avoidance to ensure the safety of the human operator is paramount. The goal of this research is the design of an efficient transportation system using a swarm of drones to expedite the production process for the factories of the future. In the context of part delivery, efficiency is defined as the accurate and timely delivery of parts, while ensuring the structural integrity of the UAVs. The objectives for this project are related to navigation, path planning and tracking, and communication with external equipment, as well as a human team. The aim of the project was to develop a proof-of-concept for a collaborative intra-logistical delivery system utilising a swarm of drones, investigating the application of decentralised control to determine an effective architecture for parts delivery. Along with this, the main objectives were as follows:

a. Investigate the perception capabilities for autonomous operation of UAVs.

b. Implement Path-Planning in a semi-structured environment.

c. Implement Real-time path following.

## Literature Review

The accurate path planning of autonomous vehicles is an important consideration to generate optimal trajectories. The length of the path, the horizontal and vertical angles, collision avoidance, height and energy efficiency are the most important constraints for the motion of UAVs. However, an additional limitation is the limited computational power of UAVs, combined with the need for real-time path planning to avoid threats. Chang L [7] proposed the use of the A* star algorithm for efficient path planning. A heuristic function is used, to evaluate the shortest distance, combined with the minimization of potential threats. The cost function was separated into the local and global optimization, considering positions in the near proximity, as well as the position of the end goal. Another approach for optimal path planning was described by Lin CL [8]. In this work, artificial potential fields were implemented. In the presence of obstacles, an artificial boundary is created, which enacts a computed repulsive force. Finally, sampling algorithms are another tool for efficient motion planning, as described by Dong Y [9]. A major limitation for these approaches is the high computational cost for dynamic environments, as well as the limited reliability in cluttered environments. The simultaneous localization and mapping of autonomous agents is important in semi-structured environments with minimum information available, enabling safe and accurate navigation. Especially in indoors environments, a global positioning system provides insufficient information and due to the limited scale and possible interference it is divided into two main tasks. Initially, the pose and position of the UAV must be estimated based on IMU measurements. The kinematics and dynamics must be considered, to estimate the state of the drones in the inertial measurement frame. However, the environment presents global constraints, due to obstacles being present. As such, the reconstruction of a map containing knowledge of the free space must be fused with the local information, for accurate control and localization. The localization of aerial vehicles depends on three highly interconnected information systems, as described by Kendoul F [10]: the optic flow, the structure from motion and the estimation of the dynamics. However, the estimation scheme requires a high bandwidth of data, as total knowledge about the drone is necessary. This presents challenges in computational efficiency. Furthermore, a small deviation in the footage can lead to inaccurate visual odometry, due to the reliance on video data. The environment presents additional constraints, introducing the need for accurate mapping. In the work of [11], the visual data is used to translate the position of obstacles in the inertial frame. While the computational complexity is reduced, as two-dimensional data is used to extract information about the three-dimensional space, two assumptions have been made. The obstacles are solely on the ground plane and remain unmoving. This can potentially limit the accuracy and efficiency of the mapping process for a non-static environment.

Accurate localisation in indoor environments, such as an industrial setting, is more challenging due to the lack of positional data from the global positioning system. Additionally, motion blur and turbulence can further decrease the stability of the dynamic system. External markers can be utilised to provide robustness and accuracy of the localisation process. In the work of [12], ArUco markers are utilised, for the precise localisation of the camera, based on the average position from different landmarks. However, high-performance onboard image processing capabilities are required, to apply the appropriate machine vision algorithm in real-time to extract the position of the corners, while also providing correction. This is achieved through adaptive thresholding, to detect square shapes through segmentation. The inner codification determines the existence of a marker. The use of swarm intelligence is especially important during the operation of multi-agent systems that require collaborative or cooperative functions. Swarm intelligence encompasses decentralised, organised, the collective behaviour of systems both natural and artificial and is commonly employed in artificial intelligence applications. The algorithms that are developed for use in swarm intelligence applications are commonly modelled on collective animal behaviour, such as flocks of birds, swarms of ants, or a hive of bees [13]. These collective animal behaviours are extremely complex and are modelled mathematically in simpler terms to allow for algorithms of cooperative behaviour to be constructed. In centralised control, a central entity is utilised for the coordination, communication, and task allocation between agents, as described by Tiderko [14]. Information such as the state, as well as the state estimation of aerial robots are imperative for collision free trajectories. Furthermore, in the context of UAV intralogistics, reliable communication is important in two cases. The first case includes large objects, which require collaborative operation for transportation. The second case includes smaller individual components, where agents are sharing a workspace. In the work of [15], a scenario of autonomous object tracking, and grasping is described. Here, a central system is utilised for failure recovery in case the leader malfunctions. However, the wireless network was insufficient due to potential packet loss, and external environmental factors. As such, the degradation into decentralised entities, that utilise perception for autonomous navigation, can assist in the robust operation of the overall system. While the communication errors in the case of centralised control are a concern, another limitation is the need for global localisation. As aerial robots are highly dynamic systems, a continuous stream of inertial measurements is necessary, which is not always feasible in centralised systems due to communication delays. As such, relative localisation based on perception capabilities can be used. Pavel Petrácek [16] utilised UVDARs, to use vision for localisation based on the position of other members of the swarm. This enabled higher scalability capabilities, due to the absence of a central communication system. Due to the increased level of autonomy, dangerous paths were avoided. Additionally, the error from the global localisation is removed. However, the inherent instability of individual agents can increase the localisation error exponentially. The position of individual drones was externally measured and used as a reference point, for the evaluation of accuracy, which is essential in dense swarms. A combination of external information, with increased individual perception, can thus lead to improved results. Finally, since a decentralised architecture does not depend on a single entity for communication, the point of failure is reduced, increasing the robustness and failure recovery capabilities of the system.

Machine vision involves the use of digital image processing methods to allow for automated perception and visual odometry. In the context of UAVs, it facilitates the use of automatic path-planning, localisation, and mapping through the manipulation of images to allow for agents to gain a semantic understanding of their environment. Through the combination of semantic environmental understanding, dynamic programming and machine-learning, robots can act through self-governing methods rather than through explicit programming. Removing the need for explicit programming greatly increases the robustness of a constructed system, allowing for it to adapt to changing and unpredictable external stimuli [17]. Through the application of machine vision, objects can be continuously tracked from a video feed, allowing for commands to be sent to an agent based on their perception of the tracked object [18]. Applying a mask to the image matrix is one method for object tracking, allowing for specific objects to be isolated based upon their 8-bit integer values in the image matrix [19]. Once the object location has been located relative to the centre of the visual feed, a moment from the object centre can be calculated and converted into velocity commands for a robotic agent, dependent on the objects X and Y offset values from the centre of the image [20]. In the case of UAVs, the tracking of an object can be used to automatically pass velocity commands to an agent, allowing for dynamic autonomous path-planning to be implemented through a master-slave relationship. The tracked object can then be attached to another autonomous or explicitly programmed agent to allow for this agent to act as the master in lieu of the object itself [21]. Object recognition involves the detection of an object, along with the classification of said object. For logistical settings, this has given increased production speeds through the application of object recognition for pick and place tasks. By applying object recognition to such tasks, systems can be made with a higher level of autonomy and robustness, expediting the programming process by removing the need for stringent programming and allowing for more dynamic methods to be utilised [22]. Through object recognition, robotic agents can gain a deeper semantic understanding of their environment, allowing for differentiation between objects which is vital for effective autonomous performance in a pick and place or transportation task. The recognition of objects also promotes the level of autonomy available to the agents, allowing for tasks to be more easily determined based upon the object that is detected. Unsupervised monocular depth estimation is a method that allows for depth estimation to be performed on a monocular image or video feed. The architectures involved in unsupervised depth estimation are often exceedingly complex and require leveraging an extremely high level of computation for real-time applications [23]. Due to the computational requirements of monocular depth estimation, the usual architecture cannot be efficiently run on simple embedded systems, such as the onboard system for a UAV. A significant portion of the computational power required is due to the decoding and encoding of the image feed.

In [24], a novel pyramidal feature-extraction architecture for monocular unsupervised depth estimation was proposed (PyD-NET) that allows for real-time performance whilst requiring significantly less computational power. This enables the up-scaling of depth maps from low-resolution decoders, requiring significantly less memory and computational power with comparable results to state-of-the-art methods. Poggi [24] compared the performance of their constructed PyD-NET architecture with the work proposed in [23], giving comparable results whilst allowing for the inference of the depth map at a rate of 2Hz using a Raspberry Pi 3. The generation of a depth map can be used in the case of UAVs to allow for autonomous and dynamic real-time collision and obstacle avoidance whilst only requiring a low-resolution video feed; however, this is severely limited by the viewing angle of the camera. Control systems can be implemented into autonomous systems to allow for the management, command, and regulation of agent's behaviour through control loops [25]. Depending on the variables being controlled, the control system can be implemented to take multiple inputs and outputs, allowing for one control system to be utilised for complex systems. This greatly simplifies the construction and tuning of the control system, with the compromise being limited individual control of the output variables [26]. Implementing a single input/output controller for each variable allows for finer control; in systems where fine control of individual variables is necessary, this can be desirable. In the case of UAVs, control systems are vital for accurate flight performance and help to stabilise the UAV during operation. The implementation of multiple single input/output PID (Proportional-Integral-Derivative) controllers, one for each plane of motion, allows for individual tuning in each direction of movement. This results in a more robust and stable system, especially when UAVs are subject to additional payloads; although requires significantly more time to tune and optimise [27].

# Methodology

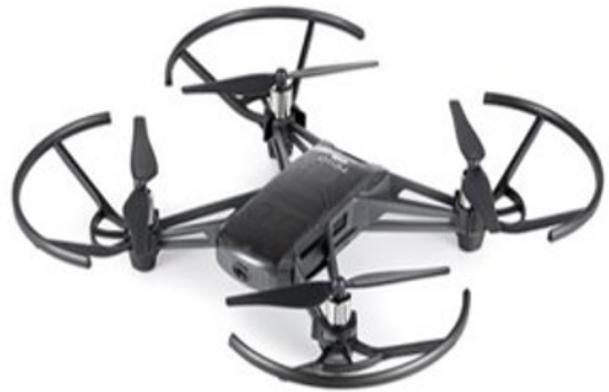

**Figure 1:** Overview of the tello drone (Tello Manual, 2018).

The UAV used in this article was the Tello EDU Drone, Figure 1. The specifications of the drone are presented in Table 1. The UAV contains a flight controller, video system, infrared positioning system, propulsion system, tilt sensor, current sensor, magnetometer, barometric pressure sensor, and flight battery. The infrared positioning system allows the drone to hover in place more precisely. Through the attachment of a mono-colour object to an agent, the object tracking/following function can be used for dynamic real-time path-planning and following. By hard-coding manual waypoints for one agent, another agent can act autonomously, following the first agent through tracking of the attached mono-colour object. This method has been utilised previously for effective real-time path-following without the need for excessive computation, as in [21]. The object to be attached to the non-autonomous agent was chosen to be a green cotton ball due to its being extremely lightweight, minimising the effects of the additional mass on the stability and control of the drone. Green was chosen as the colour for the object as the testing environment contained no objects of the same colour, reducing the risk for a false positive detection during operation. For the development of the decentralised system, two approaches were utilised, each with different architectures. The leading drone utilises the flock2 ROS2 package [28], as it can be used as a swarm controller. The package consists of four different subpackages: flock2, tello_ros [29], fiducial_vlam [30] and ros2_shared [31]. The tello_ros package was very important, as it allowed the programming of the drone through text-based commands based on the official SDK(software development kit). This was accomplished by calling the service, tello_action, which is responsible for passing a string to the onboard intelligence of the drone. It should be noted that this service cannot be interrupted. The package is comprised of two main nodes: /solo/drone_base and /solo/tello_driver. The first node represents the drone. Here, telemetry data(/solo/flight_data), as well as the state of the drone (/solo/tello_response) are received. The telemetry data includes the orientation, velocity, temperature, height, battery, barometric altitude, time of flight and acceleration. However, access to odometry is restricted. Telemetry data indicates that the connection has been successful. The response of the drone is important, as it is responsible for monitoring the battery levels, and validates whether there was a connectivity issue. However, it should be noted that this does not apply to rc commands, which directly send velocity values (x, y, z and yaw). As such, they were not utilised, to ensure safe operation. The tello_driver node is responsible for publishing the state response as well as the decoded video feedback through the /solo/image_raw and /solo/camera_info (ROS, n.d.) topics.

**Table 1:** Drone specifications.

| Weight | Approximately 80g (Propellers and Battery Included) |
|---|---|
| Dimensions | 98×92.5×41mm |
| Propellers | 3 inches |
| Built-in Function | Range Finder, Barometer. LED, Vision System, 2.4 GHz 802.11n Wi-Fi, 720p Live View |
| Port | Micro USB Charging Port |

The fiducial_vlam package was utilised for the localisation of the drones. The camera_info is used in the vloc_node, to extract the corners of the binary squares. Then, based on the inner codification, the frame of the camera is determined, based on prior knowledge of the landmarks through the fiducial_map topic. The fiducial_observations include the pose with the covariance of the camera link, based on the average position that occurs from the 8 markers. The visualisation is implemented through rviz2, which shows the movement of the camera frame, as shown in Figure 2. For the implementation of the object tracking functionality, a combination of three repositories was used: TelloCV, Tracker, and SimplePID. The TelloCV repository incorporates many functionalities that allow for the easy connection and passing of commands from a personal computer to the Tello drone, utilising the TelloPy library that enables commands to be passed to the drone for each plane of motion, allowing for fine control and individual PID tuning in each plane. Once a connection is established to the drone through TelloCV, the video feed is continuously streamed to the controlling PC, allowing for real-time manipulation and processing of the images. The Tracker functionality takes the video feed from TelloCV, encoding the UDP packets using an h264 encoder. The first step in the process is the application of Gaussian Blur to the image matrix with a kernel size of 5x5, removing excess noise from the image and aiding in the detection of contours and edges [32]. The next step involves the conversion of the image matrix data from RGB (Red-Blue-Green) colour space to HSV (Hue-Saturation-Value) colour space, this facilitates the application of a mask to the image by allowing for separation of the image dependent on the hue (wavelength) of light. A mask is then applied to the image based on a pre-defined upper and lower bound within the HSV colour space; this range is dependent on the colour of the object to be tracked. The final processing step for the image matrix is erosion then dilation. These two processes ensure that the mask is filled and does not contain any gaps that may cause errors during contour detection. The mask application converts the image into a binary occupancy map, where the object being tracked is the occupied area of the map. Contours can then be found from the binary occupancy map, using the RETR_EXTERNAL method to select only external contours. A minimum area enclosing circle is constructed around the contour, where the position of the centre of the circle relative to the image centre gives an X and Y offset value. Each plane of motion has its own PID that was implemented using the SimplePID library, with the X and Y offset values being taken as the inputs for the X and Y plane PIDs respectively. For translation of the drone in the Z plane, the radius of the minimum enclosing circle is used as the setpoint for the PID, allowing for varying sized objects to be utilised for tracking. By varying the setpoint for the Z-plane PID controller, the proximity with which the drone follows the object can be adjusted. Results from the PID controllers are taken and sent as velocity commands to the drone for their respective plane; outputting either a positive or negative result corresponding to the direction of motion in the respective plane. The output from the PID controllers is limited to be a maximum or minimum value to ensure the drone does not attempt to exceed safe operational speeds. Once all the values for the PID are calculated, the commands are passed to the drone by iterating through a constructed dictionary, allowing for dynamic motion in multiple planes simultaneously.

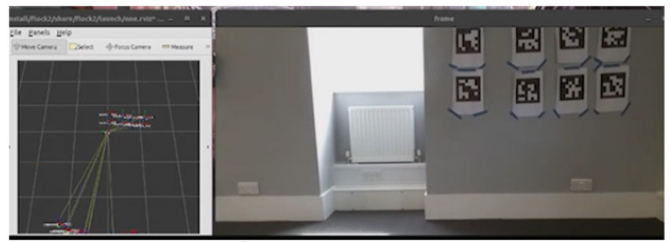

**Figure 2:** Fiducial VSLAM for experimental setup.

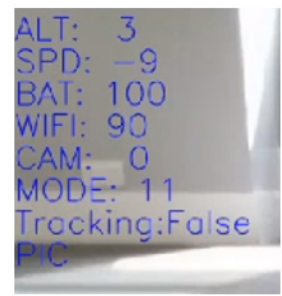

**Figure 3:** Object tracking/following HUD.

A HUD (Head-Up Display) is overlayed onto the video feed, displaying the minimum enclosing circle for the object and the moment from the object centre to the centre of the image. The HUD also displays additional information about the altitude of the drone, which is accessed by subscribing to EVENT_FLIGHT_DATA in TelloPy. An example of the HUD is shown in Figure 3. For monocular depth estimation to be performed, the PyD-NET repository was used as in [24]. This was implemented through the TelloCV repository to allow for manual control and object tracking to be integrated with the collision avoidance functionality. After the connection to the drone is established, the PyD-NET model is initiated, defining an output resolution for the depth map. A depth map is then generated which can be utilised for dynamic collision avoidance. Collision avoidance was implemented in the system, taking the generated depth map and scanning the 8-bit intensity values in the map for the maximum value. The maximum intensity value in the depth map represents the section of the image that is most proximal to the agent. If the maximum intensity value is above a set minimum intensity threshold value, the location of the maximum intensity pixel is located to allow for an X offset and Y offset value to be calculated from this pixel to the centre of the image. These X and Y offset values can be used to transmit velocity commands to the drone by iterating through a constructed dictionary of commands, resulting in the drone moving to avoid the area of highest intensity.

## Experimentation

The experimentation for the decentralised architecture involved the assessment of several individual components, as well as the testing of a semi-integrated system implemented through hardware. Two laptops were utilised with the different packages, as explained in the methodology. The experimentation that was undertaken for the object tracking/following, PyD-NET collision avoidance, infrared collision avoidance, and integrated system is outlined in this section. Several tests were performed on the object tracking/following system to optimise and iteratively improve its performance. The constructed mask was tested numerous times to determine the optimal HSV threshold values for the detection of the object. Significant testing was also performed on the PID controllers, iteratively tuning them to improve the performance of the object following functionality. As the PID controllers were implemented using single input/output controllers, the tuning time for this system was extensive and required constant tweaking of the gains of every plane of motion for optimal performance. The reaction time of the system was also tested to ensure that the system is capable of the dynamic following at relatively high speeds. Due to the unsupervised nature of the PyD-NET model, it is difficult to assess its accuracy, having no metrics or known values to compare to. This resulted in the analysis of the performance of the system being focussed more on the processing time for each frame; indicating whether the system would be usable in a real-time collision avoidance scenario. The accuracy of the depth map was determined qualitatively, by comparing the proximity of objects to the corresponding intensity values on the depth map. Testing was performed both offline with a webcam and during the operation of the drone. The experimentation undertaken on the system whilst the drone was in operation was minimal, involving the drone simply taking off, hovering for 5 seconds then landing whilst continuously monitoring the depth map outputted by PyD-NET.

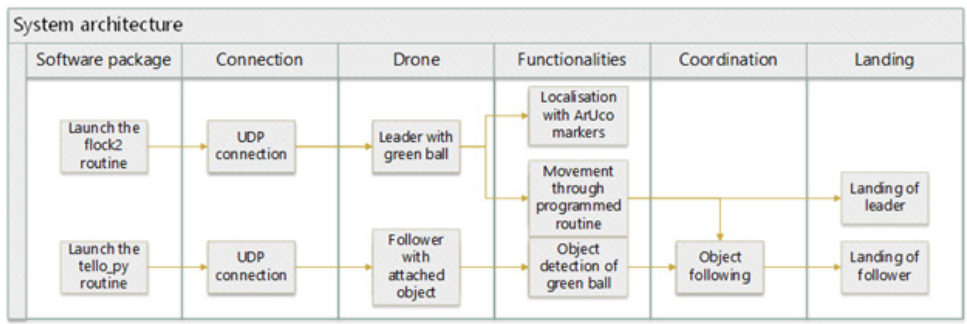

**Figure 4:** System architecture.

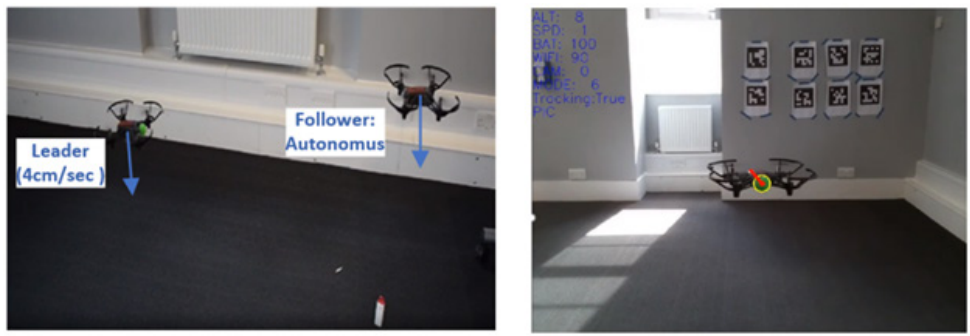

**Figure 5:** System architecture.

The decentralised integrated system consists of a master-slave relationship between the leader and the follower. The leader was manually programmed through the flock2 to follow relative waypoints. Initially, it moved 50cm upwards, for the follower to detect the object attached to it. Then, it moved forward by approximately 150cm, to ensure that collision with the wall will not occur. The speed was set to 40cm/s, for the follower drone to accurately track the position of the object and ensure smooth movement. During the operation, VSLAM(visual simultaneous localisation and mapping) localisation was used to determine the position of the drone in real-time. The localisation, while independent from the trajectory, was utilised to evaluate the performance of the ArUco markers in future development for state estimation. A green cotton ball was attached to the back of the leader drone to be used for the follower to autonomously track the position of the leader. The overall system architecture is shown in Figure 4. The following drone utilised the object tracking/following functionality to demonstrate autonomous dynamic path-following through hardware. Additionally, the following drone is carrying a small, lightweight object to prove the viability of a parts delivery system utilising drones. The stability and performance of the following drone were qualitatively assessed during operation to determine the efficacy of the object tracking/following system. The overall experimental setup and object tracking functionality are shown in Figure 5. This experimentation was used to demonstrate the viability of utilising a master-slave relationship for dynamic path-following, where the master is explicitly programmed, and the slave acts autonomously to mimic the master.

## Analysis and Results

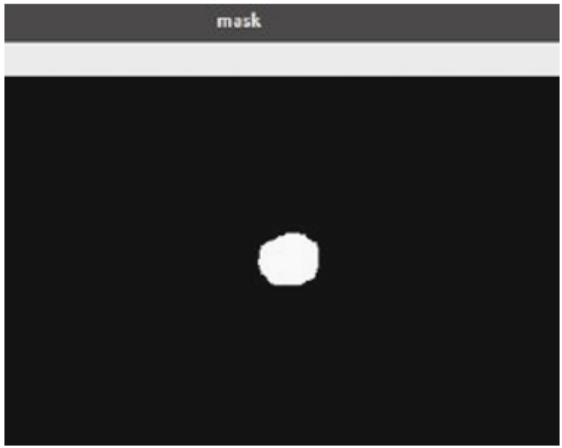

**Figure 6:** Final constructed binary occupancy map.

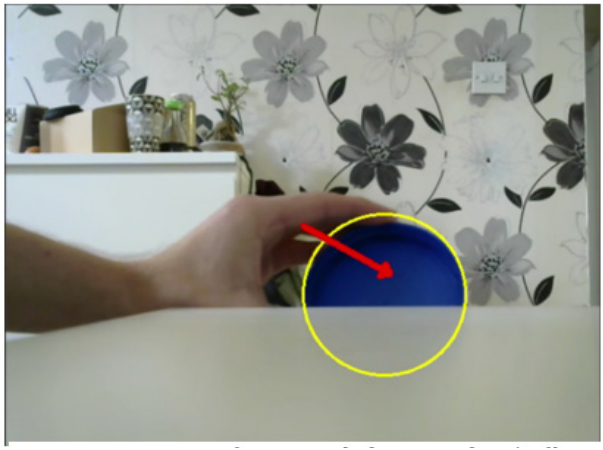

**Figure 7:** Final constructed binary occupancy map.

The mask construction for the object tracking function was iteratively improved through trial and error to ensure the mask does not capture any unwanted features in the image. For the green cotton ball used in the integrated system test, the optimal HSV minimum and maximum threshold values were found to be [40,75,20] and [80,255,255] respectively. A blue circular object was also used to assess the systems robustness and adaptability to various objects. Using these values, the mask was effectively applied to construct a binary occupancy map that only contained the desired object to be tracked, as demonstrated in Figure 6. The effective isolation of the object ensured that no unwanted contours were detected and tracked during operation, which could potentially result in a collision. A minimum enclosing circle was successfully applied to the contour of the object (shown as a yellow circle), with the moment from the object to the centre of the image being displayed by a red arrow as shown in Figure 7. This demonstrates the ability of the object tracking function to continue tracking an object after it is partially obscured. The performance of the PID controllers was determined for each plane, with smooth movement in the X and Y plane. Movement in the Z plane was smooth, although slow, potentially due to the size of the tracked object; with the set-point being set to 15 pixels for the radius of the object, movement in the forwards direction was slow as the PID was already relatively 'close' to its setpoint. This suggests that the Z-plane controller still requires some additional tuning by increasing the Kp value to increase the initial overshoot, expediting the movement in the Z-direction. The PID controllers were iteratively tuned for each plane, with the final parameters for each PID controller being shown in Table 2. The testing of the PyD-NET system gave promising results in terms of the qualitative accuracy of the generated depth map. Although only qualitative analysis could be performed at present, the depth map clearly mimics the depths in the input image even in a cluttered environment. The average processing time for each frame across three individual tests was found to be 0.231 seconds: resulting in approximately 4.3 fps (Frames per Second) when running on an Intel i5-3570K. Even using the pyramidal architecture proposed in [24], the system gave remarkably poor framerate performance which would significantly hamper its applications for a real-world industrial scenario. This poor performance on a relatively capable processor suggests that the system would have limited use on an embedded system and would result in a decreased framerate. At a framerate of 4.3 fps the drone was able to perform manoeuvres to ensure that it was facing the point of most depth, although with a significant delay in operation. During operation, it was found that the slow processing speeds and the transmission of the data in UDP packets resulted in a backlog of unprocessed frames being stored in memory. This backlog resulted in severe latency during the operation of the collision avoidance system and would need to be alleviated or removed entirely to enable its effective use. Through combining the pyramidal feature extraction architecture proposed in [24] with the lightweight encoder-decoder architecture proposed in [33] the framerate would be expected to greatly increase, as the encoding and decoding of the data is an extremely important factor in determining the speed of processing. At a framerate of 4.3 fps the drone was able to perform manoeuvres to ensure that it was facing the point of most depth, although with a significant delay in operation.

**Table 2:** Final object following PID tuning parameters.

| PID Controller | Kp | Ki | Kd |
|---|---|---|---|
| X-Plane | 0.3 | 0 | 0 |
| Y-Plane | 0.3 | 0.08 | 1 |
| Z-Plane | 0.9 | 0.06 | 0.2 |

During operation, it was found that the slow processing speeds and the transmission of the data in UDP packets resulted in a backlog of unprocessed frames being stored in memory. This backlog resulted in severe latency during the operation of the collision avoidance system and would need to be alleviated or removed entirely to enable its effective use. Through combining the pyramidal feature extraction architecture proposed in [24] with the lightweight encoder-decoder architecture proposed in [33] the framerate would be expected to greatly increase, as the encoding and decoding of the data is an extremely important factor in determining the speed of processing. Due to the computational requirements of this architecture, PyD-NET collision avoidance would only be performed on the master drone to reduce the computational load on the slave agents. Through this method, the intelligence of the slave agents can be simplified, removing the need for a hierarchical vision system. Collision avoidance based on the embedded infrared sensor removes the need for external sensors, which would increase the weight and decrease the available payload. However, the reliance on the internal system limits further development, as the distance between the drone and a surface cannot be tuned. As such, this method cannot function for distances lower than 10cm. This can be detrimental, especially in the case of dynamic obstacles, which can only be detected under specific conditions (appropriate height difference). The integrated decentralised system demonstrated promising results during operation. The master drone was consistently stable during manual waypoint following, deviating very little from the desired path. The slave drone that was utilising the object tracking/following function was effectively able to track the attached green object, as shown in Figure 7, allowing for velocity commands to effectively be passed to the drone. Movement in all planes was smooth, with a small amount of undesirable overshoot in the X-Plane during rotation; this can easily be removed through additional tuning of the Kp parameter for the X-Plane PID. Movement in the Z-Plane, although smooth, was relatively slow, resulting in the master drone gaining distance on the slave drone. This effect could be reduced by increasing the Kp parameter to increase the overshoot of the system, resulting in faster initial forward motion. The system demonstrated effective real-time autonomous path-following, implemented through a combination of hardware and software, that could be used for a pseudo master-slave relationship during the collaborative operation of drones. This would allow for one agent to perceive and map the environment whilst the other agent simply follows the attached, allowing for less computationally expensive collaborative operation. Furthermore, different levels of autonomy can be evaluated for the same system, as shown in Table 3. Additionally, a proof-of-concept testbed was constructed, which can assist future research in the domain of collaborative part delivery. As described in the work of [4], automatic operation is of great importance, to ensure predictable behavior of the aerial agents. Predictability can be important, as, during the experimentation phase, the safety of the authors in an enclosed space was a priority.

**Table 3:** Summary of autonomy levels and functionality.

| Drone | Autonomy level | Functionality |
|---|---|---|
| Leading drone | Manual programming | Path generation for following drone, VSLAM and ball for object following |
| Following drone | Full autonomy | Object following, and part delivery |

## Discussion

During the project, both the software and hardware limitations were very important considerations. For parts delivery, the low payload carrying capabilities of drones, even for collaborative lifting, limit their application to small and lightweight objects. An additional restriction was the power supply available to the drones. A detachable battery with a voltage of 3.8V and a capacity of 1.1Ah was used. This source provided an average operating time of 10 minutes. The vision, onboard intelligence and the motors are subsystems that require a high amount of energy. The application of additional perception capabilities further increases the power consumption, resulting in minimal time for testing and experimentation. The temperature also rises sharply during use due to the increased demands for onboard processing, limiting the extended use of the equipment. The need for a robust testing, validation and documentation process was vital to achieving the research objectives. The official Tello SDK was also highly restrictive. In dynamic systems, giving unrestricted access to the control software can lead to safety issues due to the need for precise tuning, especially in the case of trajectory planning and object tracking/recognition. The unofficial TelloPy library provided the capabilities for directly interfacing with the motors for the drones, but the official library was used for the leader to ensure safe trajectory tracking. The implementation of PyD-NET uncovered some limitations with the system, namely the poor framerate resulting in latency of command execution. The works of [24] and [33] could be integrated to alleviate this effect and improve the framerate of the model, thus improving its performance. Limitations were also discovered with the object tracking/following function, namely the time it takes to tune the PID for one agent. As each agent will have slightly different parameters, each agent is required to be tuned individually resulting in an extremely slow process of iterative improvements. This could be combatted by implementing on-the-fly parameter tuning for the PIDs, allowing for the dynamic altering of the control system in real-time. A multi-input/output PID controller could also be used as an alternative control system. Although this does not give the same level of robustness or flexibility as individual PID control for each plane of motion, the significant reduction in tuning time may end up being more desirable than the additional performance.

To facilitate further developments, the official Tello SDK is inadequate as only the video feedback is readily available. While TelloPy can assist through direct interfacing with the motors, it is not an ideal solution as it has limited access to the embedded hardware, thus limiting the level of control available. Specifically, as mentioned in section 5.1, visual odometry utilises kinematic, dynamic, and visual information for state estimation. The PX4 autopilot firmware is a well-documented open-source alternative flight controller [34], which enables the easier development of autonomous aerial robots. Furthermore, the PX4-ROS2 bridge, utilises a FastRTPS/DDS bridge for integration with the Robot Operating System, for more comprehensive interfacing with the hardware. Integration with ROS2 is important, as it provides a better overview of the system architecture, and assists in the sensor fusion process, through topics. Finally, the plugin suite with Gazebo (PX4, PX4 Gazebo Plugin Suite for MAVLink SITL and HITL, n.d.) can facilitate rapid software development in a controlled and safe environment. Some important capabilities include a bespoke physics-based simulator and simultaneous localisation and mapping based on different sensors. Furthermore, integration with the MAVLink protocol (MAVLINK, n.d.) better emulates the communication between the ground station and the aerial vehicles.

However, the hardware is also important. Direct access to the motors, a small size factor, and safety are important considerations. The Pixhawk 4 Mini QAV250 (Holybro, n.d.), shown in Figure 8, is an economical kit that is easy to assemble. Furthermore, it provides full access to components such as the onboard controller, the motors, and the telemetry radio. The modularity of the kit enables easy testing, as each component can be tuned, and the firmware is readily accessible. However, compared to the Tello EDU, the kit has limitations. The Tello drones used in this work have an advanced level of onboard intelligence, which handles a variety of tasks, such as visual odometry without external landmarks, and an infrared sensor for collision avoidance. The propeller guards provide additional safety, and the body of the drone can handle moderate impacts. In conclusion, a trade-off between accessibility to subsystems, software development tools, and options for embedded programming is an important consideration, to develop the framework presented in this report. The final proposed system for use in the factories of the future consists of a combination of both centralised and decentralised architecture, allowing for a dynamic system where all agents are individually intelligent. The Pixhawk 4 Mini QAV250 drone would be ideal for this use case. Through the mounting of a vision system to the ceiling of the factory, the drones could be individually localised by attaching ArUco markers to each drone. This would allow for the ceiling-mounted vision system to localise and map the individual drones through Fiducial VSLAM, with the data being processed by a central entity that distributes the constructed map to each agent. This architecture reduces the need for communication between the agents, which was determined to be a significant limitation in the collaborative operation of a swarm due to latency in message transmission and receiving. Through the application of the following, homing, cohesion, alignment and dispersion algorithms, the drones could stabilise themselves based on the relative position of the other agents in a cluster; with the relative position of each agent being transmitted to the agents through the ceiling-mounted mapping system. The leader agent would have PyD-NET collision avoidance implemented, combining the works of [24] and [33] to improve the framerate performance. Object tracking/following functionalities would also be implemented to allow for the agents to act in a pseudo master-slave relationship if required. Combining object recognition functionalities with a decentralised Swarm-GAP task allocation system, as proposed in the work of [35], would allow for the autonomous perception and determination of tasks in real-time by the agents without the need for explicit determination of tasks by a central entity. Through dynamically recognising tasks in their environment, agents can determine if they can perform a task individually or if a collaborative operation is required.

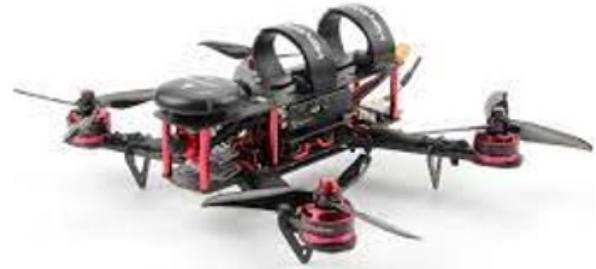

**Figure 8:** Holybro QAV250 drone.

## Conclusion

In the current work, a simple test-bed for the evaluation of different levels of autonomy was presented. The experimentation provided insight into potential capabilities in swarm robotics for parts delivery. Perception and precise path tracking can be combined through communication, which does not rely on external servers and additional sensors. Instead, vision can be used to emulate hierarchical relationships, without relying on complicated algorithms, which would decrease performance. Another important consideration is the safety of human personnel, which can be achieved through predictable behavior. Enhanced functionality can be implemented through the implementation of further cognition and swarm algorithms for bigger groups. However, on-board computational power should be a consideration, in order to maintain a balance between cognitive capabilities and smooth operation of aerial agents. It was determined that for the swarm operation of drones, taking a decentralized approach with minimal communication between agents is desirable; largely due to the delay in message transmission and receiving [36-40]. Consequentially, each agent in the swarm should be independently intelligent and have autonomous capabilities to ensure smooth collaborative operation. PyD-NET is one such solution to facilitate the onboard intelligence of the agents, allowing for real-time monocular depth estimation that can be used for autonomous navigation within an environment. When combined with localization through ArUco markers, this method would allow for real-time autonomous operation. The work undertaken could be developed further through the integration of all of the individually proposed systems, giving a coherent final solution that would enable smooth swarm operation. Additionally, further work should be done with regards to the task-allocation method to ensure the selected method requires the minimum number of messages transmitted to reduce overall delay in the system.

**For possible submissions Click below:**

Submit Article